\documentclass{article}

\usepackage{microtype}
\usepackage{graphicx}
\usepackage{booktabs} %
\usepackage{glossaries}
\usepackage{caption}
\usepackage{subcaption}
\usepackage{multirow}
\usepackage{xcolor}
\usepackage{pifont}%
\usepackage{xspace}

\usepackage{hyperref}

\usepackage[accepted]{icml2021_hacked}

\icmltitlerunning{Training Vision Transformers for Image Retrieval}

\usepackage{enumitem}
\setlist[itemize]{%
topsep=0pt,%
parsep=5pt, 
labelsep=5pt,%
labelindent=0.4\parindent,%
itemindent=0pt,%
leftmargin=*,%
itemsep=-1pt, 
}

\makeatletter
\renewcommand{\paragraph}{%
  \@startsection{paragraph}{4}%
  {\z@}{0.5em}{-1em}%
  {\normalfont\normalsize\bfseries}%
}

\begin{document}

\twocolumn[
\icmltitle{Training Vision Transformers for Image Retrieval}

\icmlsetsymbol{equal}{*}

\begin{icmlauthorlist}
\icmlauthor{Alaaeldin El-Nouby}{to,goo}
\icmlauthor{Natalia Neverova}{to} 
\icmlauthor{Ivan Laptev}{goo} 
\icmlauthor{Hervé Jégou}{to} 
\end{icmlauthorlist}

\icmlaffiliation{to}{Facebook AI}
\icmlaffiliation{goo}{ENS/Inria} 

\icmlcorrespondingauthor{Alaaeldin El-Nouby}{aelnouby@fb.com}

\icmlkeywords{Vision Transformers, Metric Learning, Image Retrieval, Zero-Shot Learning}

\vskip 0.3in]

\printAffiliationsAndNotice{}  %

\definecolor{darkgreen}{RGB}{0, 140, 0}
\definecolor{customfuchsia}{rgb}{0.77, 0.36, 0.21}
\definecolor{auburn}{rgb}{0.43, 0.21, 0.1}

\newcommand{\ignore}[1]{}

\newcommand{\cmark}{\ding{51}} \newcommand{\xmark}{\ding{55}}

\newcommand{\alaa}[1]{{\color{red}[\textbf{Alaa}: #1]}}
\newcommand{\rv}[1]{{\color{red}[\textbf{Rv}: #1]}}
\newcommand{\nn}[1]{{\color{blue}[\textbf{NN}: #1]}}
\newcommand{\ivan}[1]{{\color{green}[\textbf{Ivan}: #1]}}

\newacronym{sop}{SOP}{Stanford Online Products}
\newacronym{vit}{ViT}{Vision Transformers}
\newacronym{msa}{MSA}{Multi-Headed Self Attention}
\newacronym{ffn}{FFN}{Feed-Forward Network}
\newacronym{map}{mAP}{Mean Average Precision}
\newacronym{gem}{GeM}{Generalized Mean}
\newacronym{nca}{NCA}{Neighborhood Component Analysis}

\def \pzo {\phantom{0}}
\xspaceremoveexception{-}
\def \irto {\textbf{IRT$_{\text{O}}$}\xspace}
\def \irtf {\textbf{IRT$_{\text{L}}$}\xspace}
\def \irtr {\textbf{IRT$_{\text{R}}$}\xspace}
\def \lentropy {\mathcal{L}_{\mathrm{KoLeo}}}

\begin{abstract}

Transformers have shown outstanding results for natural language understanding and, more recently, for image classification. We here extend this work and propose a transformer-based approach for image retrieval: we adopt vision transformers for generating image descriptors and train the resulting model with a metric learning objective, which combines a contrastive loss with a differential entropy regularizer. 

Our results show consistent and significant improvements of transformers over convolution-based approaches. In particular, our method outperforms the state of the art on several public benchmarks for category-level retrieval, namely Stanford Online Product, In-Shop and CUB-200. Furthermore, our experiments on  ${\mathcal R}$Oxford and ${\mathcal R}$Paris also show that, in comparable settings, transformers are competitive for particular object retrieval, especially in the regime of short vector representations and low-resolution images.

\end{abstract}

\section{Introduction}
\label{sec:introduction}

One of the fundamental skills in reasoning is the ability to predict similarity between entities even if such entities have not been observed before. 
In the context of computer vision, 
learning similarity metric has many direct applications such as content-based image retrieval, face recognition  
and person re-identification.  
It is also a key component of many other computer vision tasks like zero-shot and few-shot learning. 
More recently, advances in metric learning have been essential to the progress of self-supervised learning,  
which relies on matching two images up to data augmentation as a learning paradigm. 

Modern methods for image retrieval typically rely on convolutional encoders and extract compact image-level descriptors. Some early approaches used activations provided by off-the-shelf pre-trained models \cite{babenko2014neural}. 
However, models trained specifically for the image retrieval task achieve better performance  \cite{radenovic2018fine, teh2020proxynca, wang2020cross}. A number of efficient objective functions have been proposed to penalize the discrepancy between computed similarities and the ground truth. In addition, research has been focused on improvements of sampling methods and data augmentation strategies. 

\begin{figure}[t!]
\centering
\includegraphics[width=\linewidth]{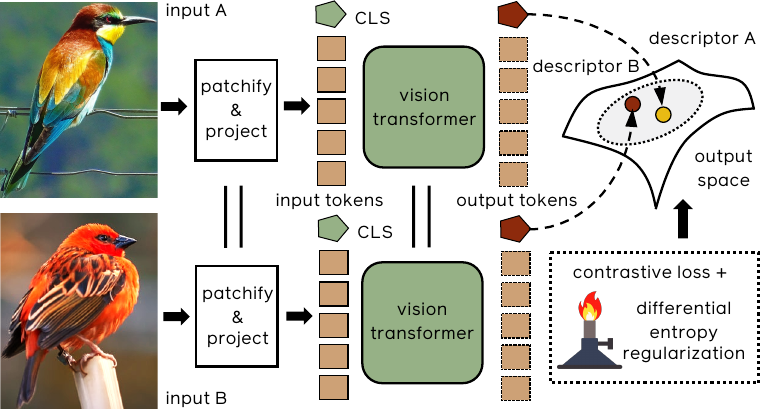} %
\vspace*{-1.0\baselineskip}
\caption{We train a transformer model with a Siamese architecture for image retrieval. Two input images are mapped by the transformers into a common feature space. At training time, the contrastive loss is augmented with an entropy regularizer.   
\label{fig:introfig}
}
\vspace*{1\baselineskip}
\end{figure}

The transformer architecture by \citet{vaswani2017attention} has been successfully used for a number of NLP tasks~\cite{devlin2018bert, radford2018improving}, and more recently in the core computer vision task of image classification \cite{dosovitskiy2020image, touvron2020deit}. This is an interesting
development as transformer-based models adapted for computer vision come with a different set of
inductive biases compared to the  currently dominant convolutional architectures. 
This suggests that such models may find alternative solutions and avoid errors that are typical for convolutional backbones.
While there have been some efforts exploring attention-based metric learning for images 
\cite{kim2018attention,chen2019hybrid}, to our knowledge the adoption of a plain transformer has not been studied in this context. 

In this work, we introduce and study \textit{Image Retrieval Transformers} (IRT). As illustrated in Figure~\ref{fig:introfig}, our IRT model instantiates a Siamese architecture with a transformer backbone.
We investigate the adaptation of metric learning techniques and evaluate how they interplay with transformers. 
In particular, we adopt a contrastive loss \cite{hadsell2006dimensionality}, which has recently been reaffirmed as a very effective metric learning 
objective \cite{musgrave2020metric, wang2020cross}. We also employ a differential entropy regularization that favors uniformity over the representation space
and improves performance.

We perform an extensive experimental evaluation and validate our approach by considering two image retrieval tasks. First, we investigate the task of category-level image retrieval, which is often used to measure the progress in metric learning \cite{teh2020proxynca, musgrave2020metric}.
We also explore retrieval of particular objects, and compare our method to convolutional baselines in similar settings (same resolution and similar complexity).

 The main contributions of this work are listed below. 

\begin{itemize}
    \item We propose a simple way to train vision transformers both for category-based level and particular object retrieval, and achieve competitive performance when compared to convolutional models with similar capacity.
    \item As a result, we establish the new state of the art on three popular benchmarks for category-level retrieval. %

    \item For particular object retrieval, in the regime of short-vector representation (128 components), our results on ${\mathcal R}$Oxford and ${\mathcal R}$Paris are comparable to those of convnets operating at a much higher resolution and FLOPS. 
    \item We show that the differential entropy regularizer enhances the %
    contrastive loss and improves the performance overall. 
\end{itemize}
\medskip

\section{Related Work}
\label{sec:related}

\paragraph{Transformers.} The transformer architecture was introduced by \citet{vaswani2017attention}
for machine translation. %
It solely relies
on self-attention and fully-connected layers, and achieving an attractive trade-off between efficiency
and performance. 
It has subsequently provided state-of-the-art performance
for several NLP tasks~\cite{devlin2018bert, radford2018improving}. %
In computer vision, several attempts have been devoted to incorporate various forms of attention, 
for instance in conjunction \cite{wang2018non} or as a replacement to convolution \cite{ramachandran2019stand}. 
Other methods utilize transformer layers on top of convolutional trunks \cite{carion2020end} for detection.

More
recently, convolution-free models that only rely on transformer layers have shown competitive
performance \cite{chen2020generative, dosovitskiy2020image,touvron2020deit}, positioning it as a
possible alternative to convolutional architectures. In particular, the \gls{vit} model proposed by
\citet{dosovitskiy2020image} is the first example of a transformer-based method to match or even surpass state-of-the-art convolutional models on the task of image classification. 
\citet{touvron2020deit} subsequently improved the optimization procedure, leading to competitive results with ImageNet-only training~\cite{deng2009imagenet}.

\paragraph{Metric Learning.} 
A first class of deep metric learning methods 
is based on classification: these approaches represent each category using one
\cite{movshovitz2017no,teh2020proxynca, zhai2018classification, boudiaf2020unifying} or multiple
 prototypes \cite{qian2019softtriple}. The similarity and dissimilarity training signal is computed against the prototypes rather than between individual instances. Another class of methods operate on pairs methods: the training signal is defined by  %
similarity/dissimilarity between individual instances
directly. A contrastive loss \cite{hadsell2006dimensionality} aims to push representations of
positive pairs closer together, while representations of negative pairs %
are encouraged to have larger distance. The triplet loss
\cite{weinberger2009distance} builds on the same idea but requires the positive pair to be closer
than a negative pair by a fixed margin given the same anchor. \citet{wu2017sampling} %
proposes negative sampling weighted by pair-wise distance to emphasize harder negative examples. Other
pair-based losses rely on the softmax function \cite{NIPS2004_42fe8808, sohn2016improved,
wu2018improving, wang2019multi}, allowing for more comparisons between different positive and negative pairs.

While a vanilla contrastive loss has
been regarded to have a weaker performance when compared to its successors like triplet
and margin \cite{wu2017sampling} losses, recent  efforts
\cite{musgrave2020metric} showed that a careful implementation of the contrastive loss leads to
 results outperforming many more sophisticated losses.
Additionally, \citet{wang2020cross} showed that when augmented with an external memory to allow sampling of a sufficient number of hard negatives, contrastive loss achieves a state-of-the-art performance on multiple image retrieval benchmarks.

\paragraph{Particular Image Retrieval} has progressively evolved from methods based on local descriptors to convolutional encoders.  
In this context, an important design choice is how to compress the spatial feature maps of activations into a vector-shaped descriptor%
~\cite{babenko2015aggregating,tolias2015particular}. Subsequent works have adopted end-to-end training \cite{gordo2016deep,radenovic2018fine, revaud2019learning} with various forms of supervision. In a concurrent work, \citet{gkelios2021investigating} investigated off-the-shelf pre-trained ViT models for particular image retrieval.

\paragraph{Differential Entropy Regularization.} 
\citet{zhang2017learning} aim a better utilization of the space by spreading out the descriptors through matching first and second moments of non-matching pairs with points uniformly sampled on the sphere. 
\citet{wang2020understanding} provide a
theoretical analysis for contrastive representation learning in terms of alignment and uniformity on
the hypersphere. In the context of face recognition, \citet{duan2019uniformface} argue for spreading
the class centers uniformly in the manifold, while \citet{zhao2019regularface} minimize the angle
between a class center and its nearest neighbor in order to improve inter-class separability. 

We focus our study on pairwise losses, the contrastive loss in particular, aiming to
prevent the collapse in dimensions that happens as a byproduct of adopting such an objective. Most
related to out method, \citet{sablayrolles2018spreading} propose a differential entropy
regularization based on the estimator by \citet{kozachenko1987sample}, in order to spread the vectors
on the hypersphere more uniformly, such that it enables improved lattice-based quantization
properties. \citet{bell2020groknet} adopted it as an efficient way to binarize features output by convnets in commerce applications. 

\section{Methods}
\label{sec:methods}

In this section, after reviewing the transformer architecture, we detail how we adapt it to the category-level and particular object retrieval. Note, in the literature these tasks have been tackled by distinct techniques. In our case we use the same approach for both of these problems. We gradually introduce its different components, as follows:
\begin{itemize}
    \item \irto -- off-the-shelf extraction of features from a ViT backbone, pre-trained on ImageNet; %
    \item \irtf -- fine-tuning a transformer with metric learning, in particular with a contrastive loss; %
    \item \irtr -- additionally regularizing the output feature space to encourage uniformity. 
\end{itemize}

\subsection{Preliminaries: Vision Transformer}
Let us review the main building blocks for
transformer-based models, and more specifically of the recently proposed \gls{vit} architecture by \citet{dosovitskiy2020image}. 
The input image is first decomposed into $M$ fixed-sized patches (e.g.
16$\times$16). Each patch is linearly projected into $M$ vector-shaped tokens and used as an input
to the transformer in a permutation-invariant manner. The location prior is incorporated by adding a learnable 1-D
positional encoding vector to the input tokens. An extra learnable CLS token is added to the
input sequence such that its corresponding output token serves as a global image representation.

The transformer consists of $L$ layers, each of which is composed of two main blocks: a
\gls{msa} layer, which applies a self-attention operation to different projections of the
input tokens, and a \gls{ffn}. Both the \gls{msa} and \gls{ffn} layers are preceded by layer
normalization and followed by a skip connection. We refer the reader to \citet{dosovitskiy2020image}
for details.

\paragraph{Architectures.} 

Table~\ref{tab:param_count} presents the neural networks models used through this paper. They are all pre-trained on ImageNet1k \cite{deng2009imagenet} only. 
In order to have a fair comparison with other retrieval methods, we choose to use the DeiT-Small variant of the \gls{vit} architecture introduced by \citet{touvron2020deit} as our primary model. 
The DeiT-Small model has
a relatively compact size which makes it comparable to the widely adopted ResNet-50 convolutional
model \cite{he2016deep} in terms of parameters count and FLOPS, as shown in Table~\ref{tab:param_count}. Additionally, we 
provide some analysis and results of larger models like ResNet-101 and DeiT-Base, as well as DeiT variants with advanced pre-training. 

\begin{table}[t]
\vspace{-5pt}
    \caption{Parameters count, FLOPS and ImageNet Top-1 accuracy for convolutional baselines ResNet-50 (R50) and ResNet-101 (R101) at resolution  224x224, as well as 
    transformer-based models: DeiT-Small (DeiT-S) and DeiT-Base (DeiT-B)  \cite{touvron2020deit}. \textit{$\dagger$:
    Models pre-trained with distillation with a convnet trained on ImageNet1k.}}
    \begin{center}
    \scalebox{0.8}{
        \begin{tabular}{lccc}
        \toprule
         Model& 
         \# params & FLOPS (G) & Top-1 (\%) \\
         \midrule
         R50 & 23M  & 8.3  & 76.2 \\
         DeiT-S & 22M  & 8.4  & \textbf{79.8} \\
         DeiT-S$\dagger$ & 22M  & 8.5  & \textbf{81.1} \\
         \midrule
         R101 & 46M & 15.7  & 77.4 \\
         DeiT-B & 87M & 33.7  &  \textbf{81.8} \\
         DeiT-B$\dagger$ & 87M & 33.8  &\textbf{83.9} \\
         \bottomrule
    \end{tabular}}
    \end{center}
    \label{tab:param_count}
    \vspace*{-\baselineskip}
\end{table}

\subsection{\textbf{IRT$_{\text{O}}$:} off-the-shelf features with Transformers} 
\label{sec:off-shelf}

We first consider the naive approach \irto, where we extract features directly from a transformer pre-trained on ImageNet. This strategy is in line with early works on image retrieval with convolutional networks \cite{babenko2014neural},  which were featurizing activations. 

\paragraph{Pooling.} 
We extract a compact vector descriptor that represents the image globally. In the ViT architecture, 
pre-classification layers output $M+1$ vectors corresponding to $M$ input patches and a class (CLS) embedding. 

In our referent pooling approach, \textbf{CLS}, we follow the spirit of BERT \cite{devlin2018bert} and ViT models, and view this class embedding as a global image descriptor. 
In addition, we investigate performance of global pooling methods that are typically used by convolutional metric
learning models, including average, maximum and \gls{gem} pooling, and apply them to the $M$ output tokens.

\paragraph{$l_{2}$-Normalization and Dimensionality Reduction.} We follow the common practice of projecting the descriptor vector into a unit ball after pooling. 
In the case when the target dimensionality is smaller than that provided by the architecture, we optionally reduce the vector by principal component analysis (PCA) before normalizing it.

\subsection{\irtf: Learning the Metric for Image Retrieval}

We now consider a metric learning approach for image retrieval, denoted by \irtf. It is the dominant approach to both category-level and particular object retrieval. In our case we combine it with transformers instead of convolutional neural networks. 
We adopt the contrastive loss with cross-batch memory by \citet{wang2020cross} and fix the  margin $\beta$\,=\,$0.5$ by default for our metric learning objective. 

The contrastive loss maximizes the similarity between encoded
low-dimensional representations $z_i$ of samples with the same label $y$ (or any other pre-defined
similarity rule). Simultaneously, it minimizes the similarity between representations of samples
with unmatched labels which are referred to as negatives. For the contrastive loss, only negative
pairs with a similarity higher than a constant margin $\beta$ contribute to the loss. This %
prevents the training
signal from being overwhelmed by easy negatives. Formally, the contrastive loss over a batch of size $N$ is defined as:
\begin{equation}
    \footnotesize
    \mathcal{L}_{\text{contr.}}\! = \!\frac{1}{N} \sum_{i}^{N}\!\left[\sum_{j:y_{i}=y_{j}}\!\!\!
    \left[1 - z_{i}^{T}z_{j}\right] + \!\!\!\!\sum_{j:y_{i} \neq y_{j}}\!\!\!\left[z_{i}^{T}z_{j} - \beta\right]_{+}\!\right]\!\!.
\end{equation}
The representations $z_i$ are assumed to be $l_{2}$-normalized, therefore the inner product is equivalent
to cosine similarity.

\subsection{\irtr: Differential Entropy Regularization}

Recently, \cite{boudiaf2020unifying} studied connections between a group of pairwise losses and
maximization of mutual information between  learned representations~$Z=\{z_i\}$ and corresponding ground-truth labels~$Y=\{y_i\}$. We are interested in the particular case of the contrastive loss. The mutual information is
defined as
\begin{equation}
\vspace*{-3pt}
    \footnotesize
    \mathcal{I}(Z, Y) = \mathcal{H}(Z) - \mathcal{H}(Z | Y).
\end{equation}
The positive term of the contrastive loss leads to minimization of the conditional
differential entropy $\mathcal{H}(Z | Y)$, where intuitively, samples representations belonging to the same category are
trained to be more similar:
\vspace*{-3pt}
\begin{equation}
    \footnotesize
    \mathcal{H}(Z | Y) \propto  \frac{1}{N} \sum_{i}^{N} \sum_{j:y_{i}=y_{j}} \left[1 - z_{i}^{T}z_{j}\right].
\end{equation}
On the other hand, the negative term of this loss is responsible for preventing trivial solutions
where all sample representations are collapsed to a single point. Therefore, it maximizes the entropy of the learned representations:
\begin{equation}
\vspace*{-3pt}
\label{eq:entropy}
    \footnotesize
    \mathcal{H}(Z) \propto - \frac{1}{N} \sum_{i}^{N} \sum_{j:y_{i} \neq y_{j}} [z_{i}^{T}z_{j} - \beta]_{+}.
\end{equation}
The margin $\beta$ plays an important role in the training dynamics. Low values of
$\beta$ allow exploration of a larger number of negative samples. Yet in this case %
easy negatives can dominate the training and cause the performance to plateau. In contrast, higher values of $\beta$ would only accept hard negatives, possibly leading to noisy gradients and unstable training \cite{wu2017sampling}. 

\paragraph{Our regularizer.} Motivated by the entropy maximization view of the negative contrastive term
in Equation~\ref{eq:entropy}, we add an entropy maximization term that is independent of
the negative samples accepted by the margin. In particular, we use the differential entropy loss
 proposed by \citet{sablayrolles2018spreading}. It is based on the \citet{kozachenko1987sample} differential entropy estimator:\vspace*{-5pt}
\begin{equation}
    \footnotesize
    \lentropy = - \frac{1}{N} \sum_{i}^{N} \log(\rho_{i}),
\end{equation}
where %
$\rho_{i}=\text{min}_{i\neq j}\|z_{i} - z_{j}\|$.
In other words, this regularization maximizes the distance between every point and its nearest neighbor, and therefore alleviates the collapse issue. %
We simply add the regularization
term to the contrastive loss weighted by a regularization strength coefficient $\lambda$: $\mathcal{L} = \mathcal{L}_{\text{contr.}} + \lambda \lentropy$.

Intuitively, the different entropy
regularization prevents the representations of different samples from %
lying too close on the hypersphere,
by increasing their distance from positive examples, and the hard negatives as well. Having hard negatives
with extremely small distances is a main source of noise in the training signal, as identified by
\citet{wu2017sampling}.

\begin{figure*}[t!]

    \begin{minipage}[t]{0.32\textwidth}
    \includegraphics[trim=5 0 0 0, clip, scale=0.36]{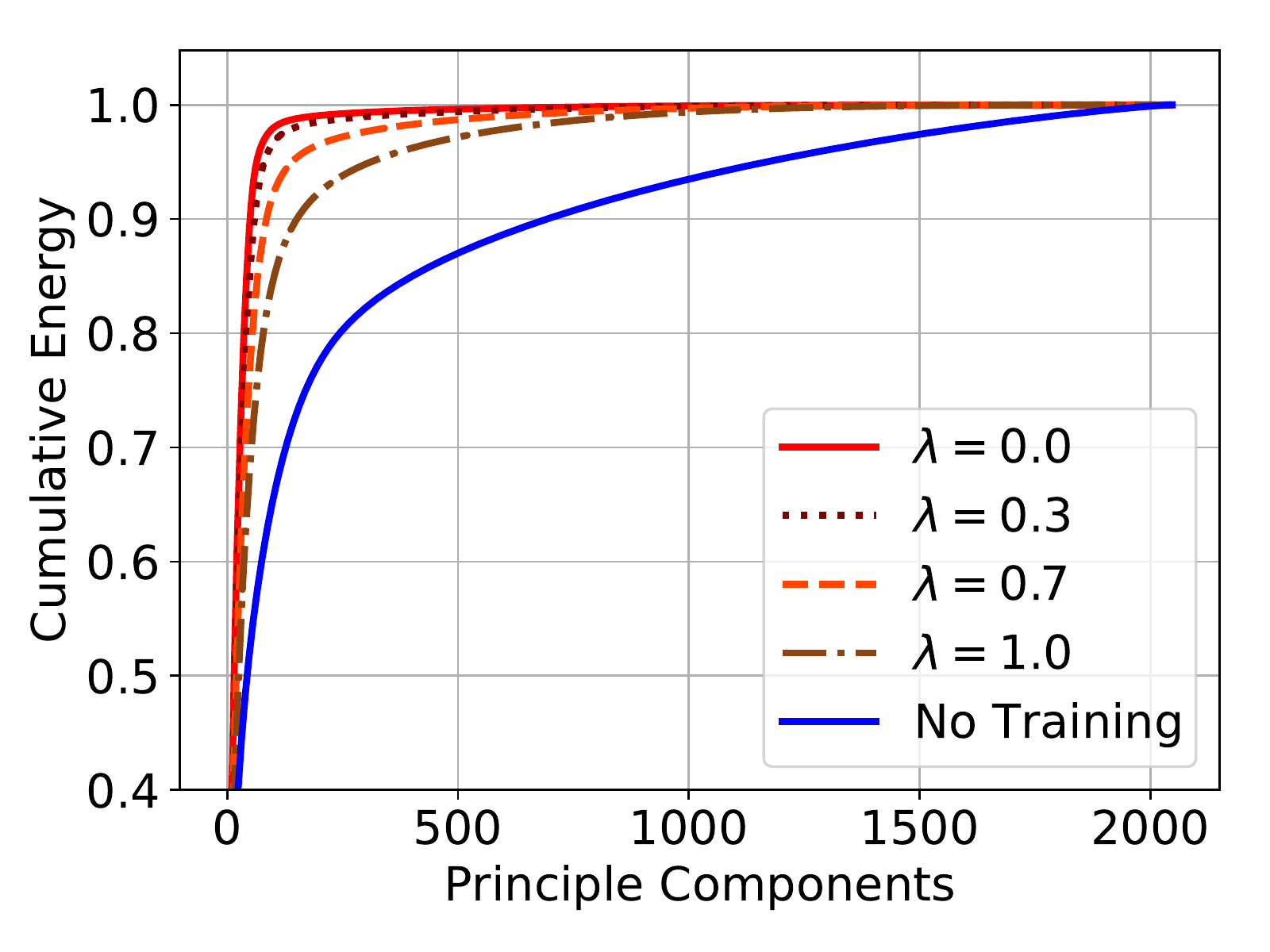}
    \vspace{-0.5cm}
    \scriptsize
    \caption{Cumulative energy of the principle components for features extracted using a ResNet-50
    backbone from the SOP dataset, with pre-training on ImageNet, with (red) or without (blue) finetuning. The
    solid red line indicates the vanilla contrastive loss with $\beta=0.5$. The features have collapsed to few dimensions after training, but the collapse is reduced by entropy regularization.
    }
    \label{fig:pca_energy}
    \end{minipage}
    \hfill
    \begin{minipage}[t]{0.32\textwidth}
    \includegraphics[trim=5 0 0 0, clip, scale=0.36]{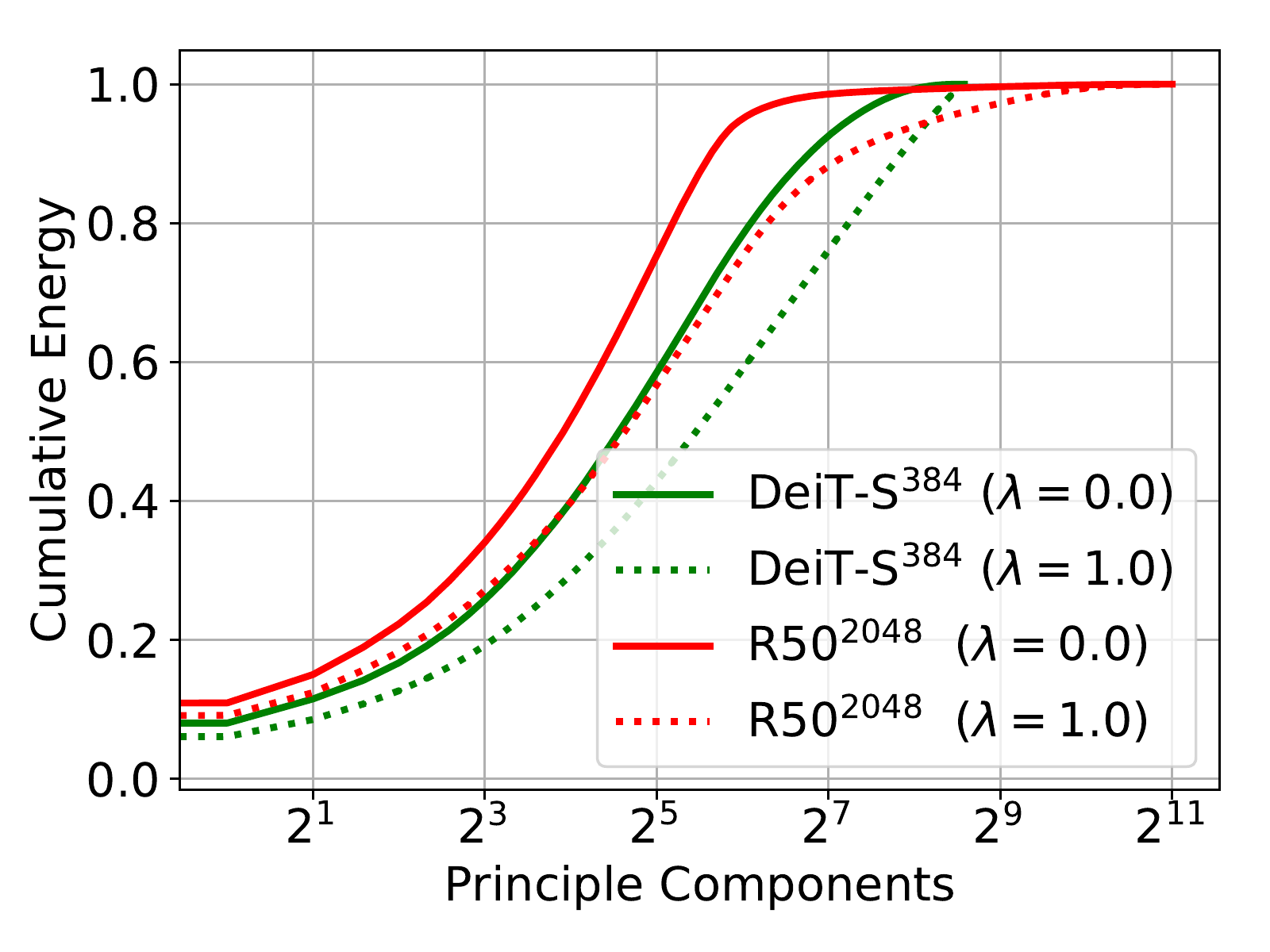}
    \vspace{-0.5cm}
    \scriptsize
    \caption{Cumulative energy of the principle components for ResNet-50 and DeiT-Small models trained using a margin of $\beta=0.5$. We can see that despite the lower extrinsic dimensionality of DeiT-S ($384 \approx 2^{8.5}$), it has higher intrinsic dimensionalities than ResNet-50 after training. This suggests that the transformer-based architectures can be more robust against the feature collapse.}
    \label{fig:pca_deit_res}
    \end{minipage}
    \hfill
    \begin{minipage}[t]{0.32\textwidth}
        \includegraphics[trim=5 0 0 0, clip,  scale=0.36]{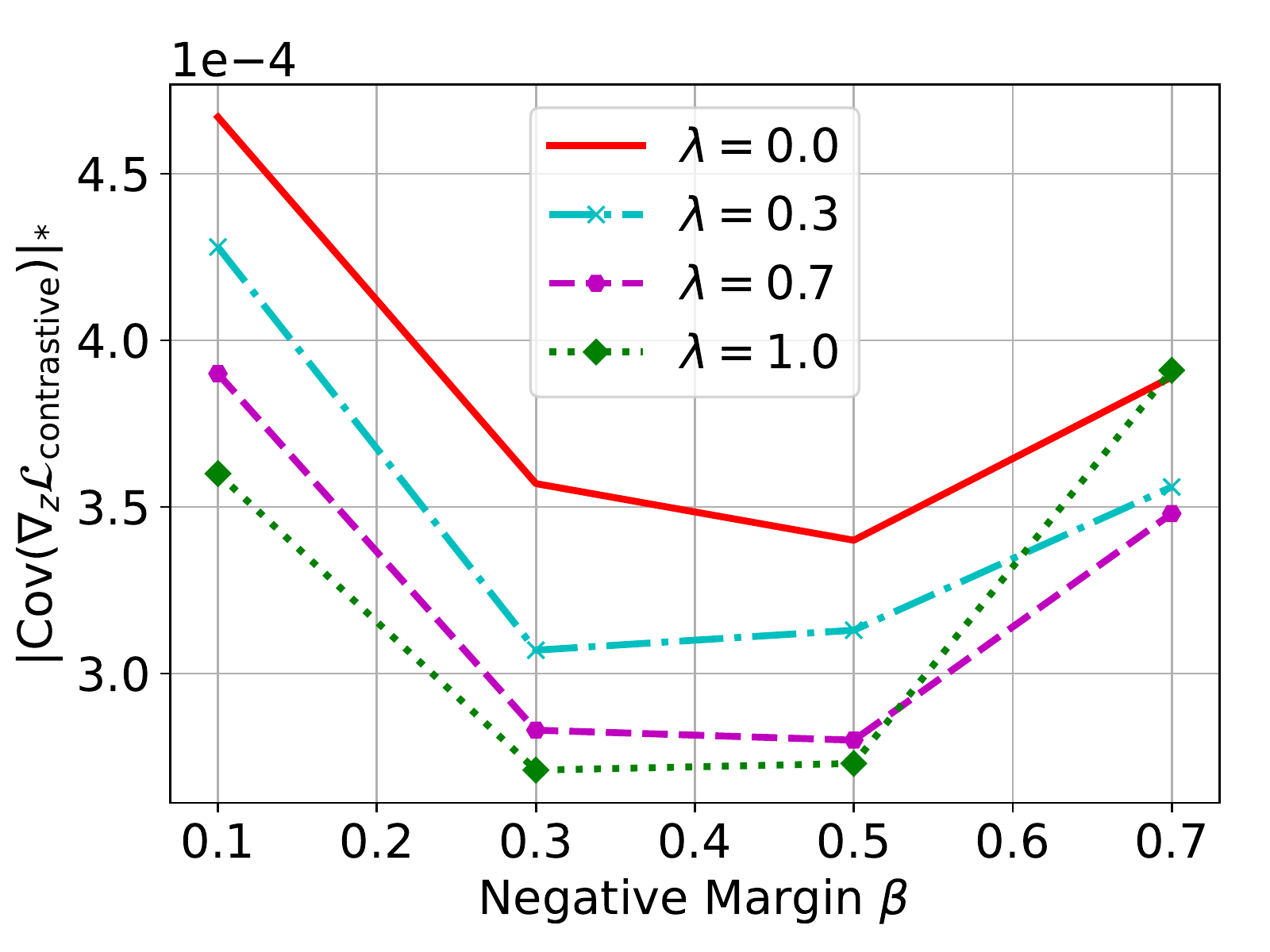}
        \scriptsize
        \vspace{-0.5cm}
        \caption{The nuclear norm for the covariance of the gradient direction, computed for different margin values $\beta$ and entropy regularization strengths $\lambda$. The gradient signal is noisier for very high and very low margin values. The $\lentropy$ regularization provides a more stable gradient signal, but, in turn, becomes harmful with high values of $\lambda$ (e.g. $\lambda=1.0, \beta=0.7$).}
        \label{fig:nuclear_cov}

\end{minipage}
\vspace*{0.3cm}
\end{figure*}

\subsection{Analysis}
\label{sec:analysis}

We study the behaviour of the output representation space when training with a contrastive loss, and how augmenting this loss with a differential entropy regularization impacts the space properties and the model performance.

\paragraph{PCA in the Embedding Space.} 
In Figure~\ref{fig:pca_energy} we examine the cumulative energy of the principle components for
features from an off-the-shelf, ImageNet pre-trained model, as well as models trained using
contrastive loss. 
We observe that the features after training with the contrastive loss suffer from a collapse in dimensions compared to an untrained model. 
This suggests an ineffective use of the representational capacity of the embedding space, as alignment is favored over uniformity while both are necessary for good representations \cite{wang2020understanding}. As we augment the contrastive loss with the differential entropy regularization,  
the cumulative energy spreads across more dimensions (see Figure~\ref{fig:pca_energy} with non-zero values of $\lambda$). Higher values of $\lambda$ alleviate  the dimensionality collapse problem. 

Another observation is that the transformer-based architecture is less impacted than convnets by the collapse (see Figure~\ref{fig:pca_deit_res}). Despite having a lower extrinsic dimensionality compared to the ResNet-50 model, the DeiT-Small features are more spread over principle components. A possible reason for that behavior is that in multi-headed attention,  each input feature is projected to different sub-spaces before the attention operation, reducing the risk of collapse.

\paragraph{Gradient Analysis.} 
As pointed out by \citet{wu2017sampling}, very
hard negatives can lead to noisy gradients. We examine the nuclear norm $\|\cdot\|_*$ associated with the covariance matrix of
the gradients directions $\gamma =\|\text{Cov}(\nabla_{z}\mathcal{L_{\text{contr.}}})\|_{*}$, 
averaged over all training iterations (see  Figure~\ref{fig:nuclear_cov}). Higher values of $\gamma$ could indicate
noisy gradients. We observe them for both very high and very low 
values of margin $\beta$ which aligns with our understanding that very easy and very hard negatives lead to less
informative and less stable training signal. Moreover, we observe a decrease in the $\gamma$ values
after the addition of the entropy regularization term.

\begin{figure*}[t!]
    \begin{subfigure}[t]{0.3\textwidth}
    \centering
    \includegraphics[trim=40 0 30 30, clip, scale=0.38]{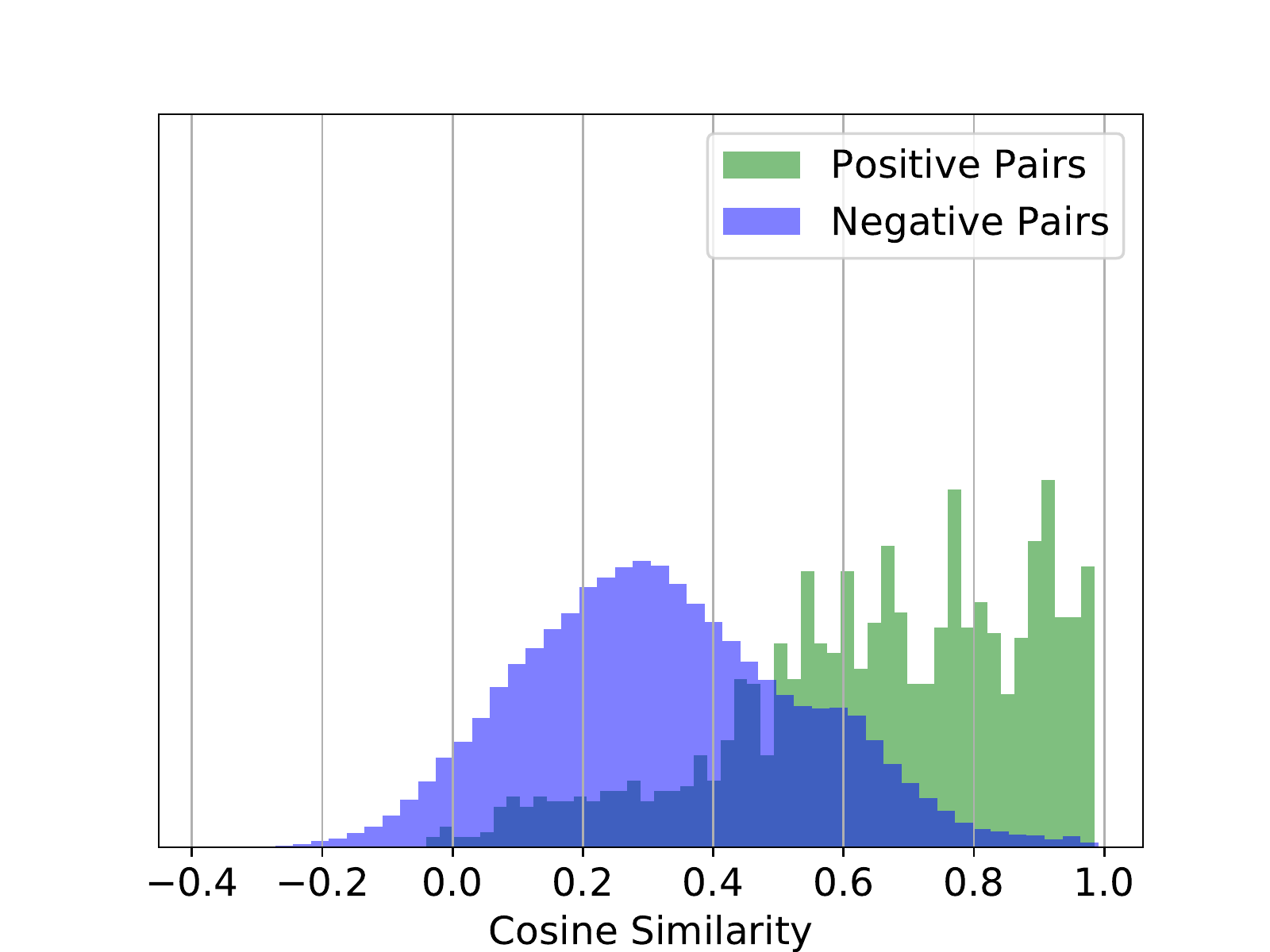}
    \scriptsize
    \caption{\irto: off-the-shelf descriptors. }
    \label{fig:hist_deit_off}
    \end{subfigure}
    \hfill
    \begin{subfigure}[t]{0.3\textwidth}
    \centering
    \includegraphics[trim=40 0 30 30, clip, scale=0.38]{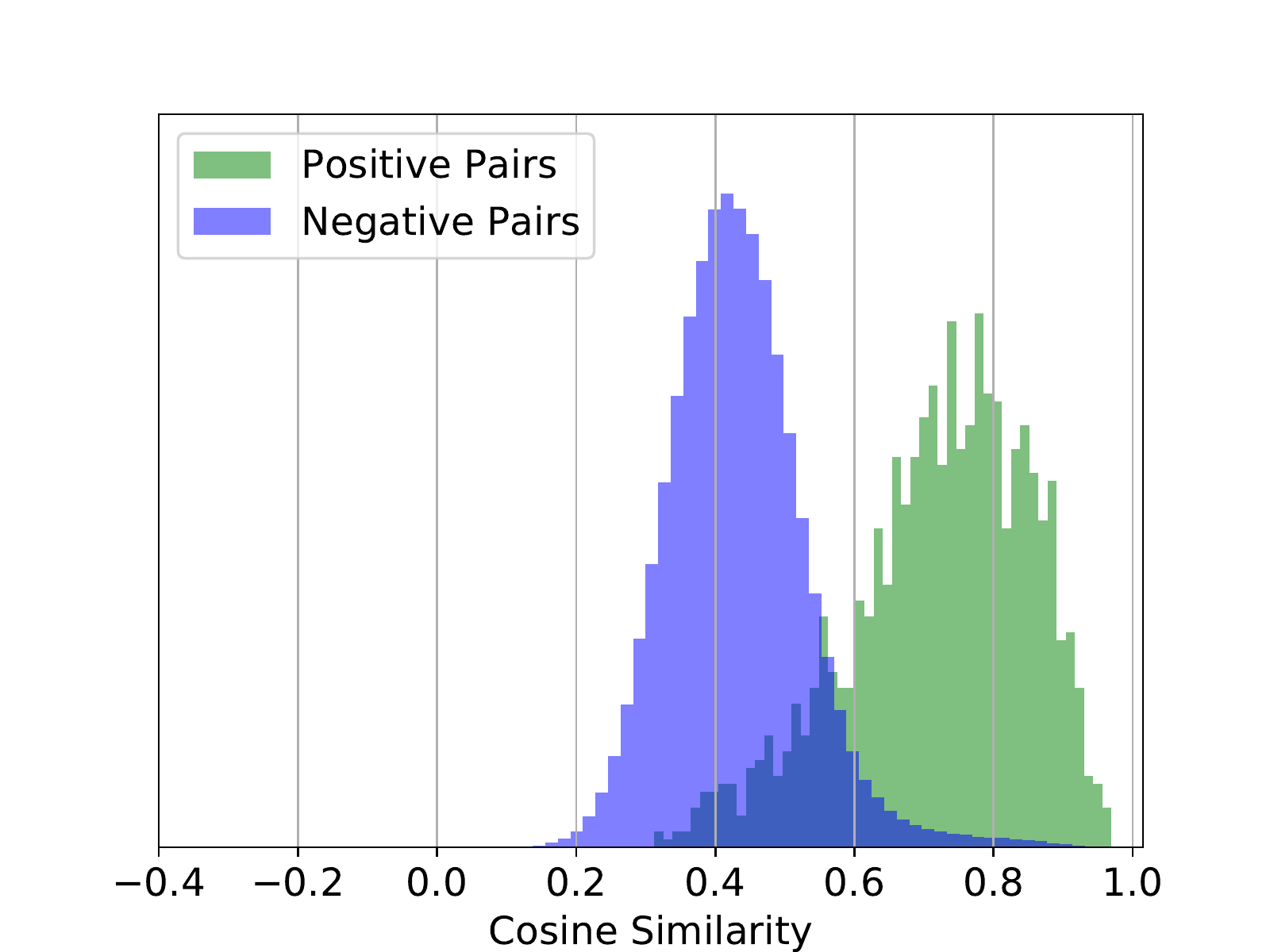}
    \scriptsize
    \caption{\irtf: finetuned descriptors.}
    \end{subfigure}
    \hfill
    \begin{subfigure}[t]{0.3\textwidth}
    \centering
    \includegraphics[trim=40 0 30 30, clip, scale=0.38]{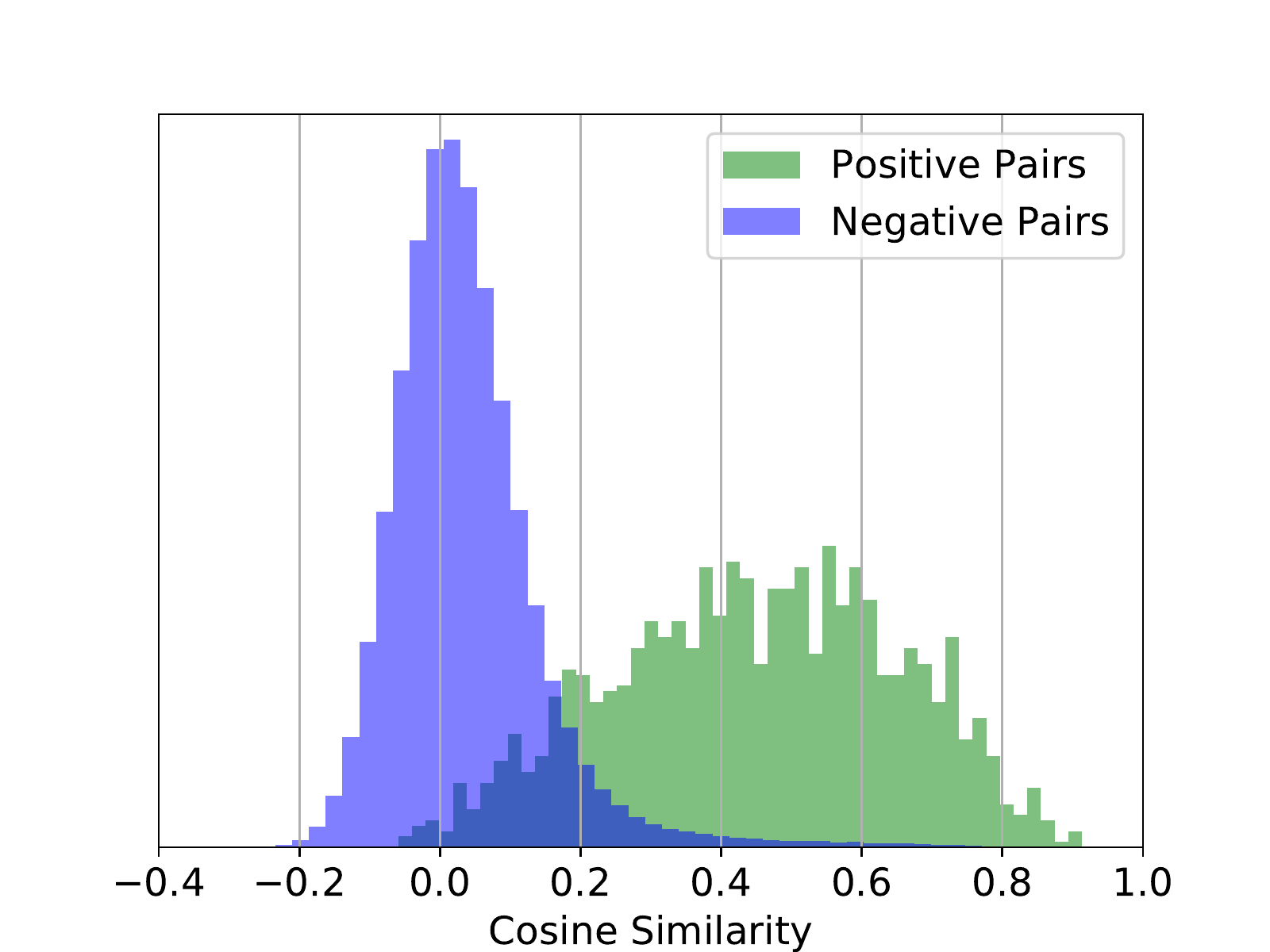}
    \scriptsize
    \caption{\irtr: finetuned with $\lambda=4.0$.}%
    \label{fig:hist_res}
    \end{subfigure}
    \caption{We present histograms for the cosine similarities belonging to positive and negative pairs for three DeiT-Small variants. The descriptors are extracted from the $\mathcal{R}$Paris (M) dataset with input images of size 224$\times$224 and descriptor dimensionality of 256-d for all models. Extracting features using off-the-shelf features (left) results in a highly overlapping positive and negative distribution.  Finetuning the model using contrastive loss (middle) reduces the overlap significantly. However, the descriptors concentrate relatively close to each other, not taking full advantage of the embedding space. Augmenting the contrastive loss with entropy regularization (right) results in a more uniformly spread distribution of descriptors and a better separation of positive and negative pairs based on absolute similarities.}
    \label{fig:hist_deit}
\end{figure*}

\section{Experiments \& Ablation Studies}
\label{sec:results}

We first describe datasets and implementation details, and then proceed with discussions of empirical results.

\subsection{Datasets}
\paragraph{Category-level Retrieval.
}

We report performance on three popular datasets commonly used for category-level retrieval. %
\textbf{Stanford Online Products}  (SOP) \cite{oh2016deep} consists of online
products images representing 22,634 categories. Following the split proposed by~\citet{oh2016deep}, we use first 11,318 categories for training and the remaining 11,316 for testing. 
\textbf{CUB-200-2011} \cite{wah2011caltech} contains 11,788 images corresponding to 200 bird categories. Following  \citet{wah2011caltech}, we split this dataset into two class-disjoint sets,  each with 100 categories for training and testing.
\textbf{In-Shop} \cite{liuLQWTcvpr16DeepFashion} contains 72,712 images of clothing items
belonging to 7,986 categories, 3,997 of which used for training. The remaining 3,985 categories are split into 14,218  query and 12,612 gallery images for testing.
We compute the Recall@K evaluation metric for a direct comparison with previous methods.

\paragraph{Particular Object Retrieval.}

For training, we use the \textbf{SFM120k} dataset \cite{radenovic2018fine} which is obtained by applying structure-from-motion and 3D reconstruction to large unlabelled image collections \cite{schonberger2015single}. The positive images are selected such that enough 3D points are co-observed with the query image, while negative images come from different 3D models. We use 551 3D models for training and 162 for validation.

For evaluation, we report results using revisited benchmarks \cite{radenovic2018revisiting} of the \textbf{Oxford} and \textbf{Paris}~\cite{philbin2007object, philbin2008lost} datasets. These two datasets each contain 70 query images depicting buildings, and additionally include 4993 and 6322 images respectively in which the same query buildings may appear. The revisited benchmarks contain 3 splits: Easy (E), Medium (M) and Hard (H), grouped by gradual difficulty of query/database pairs. (E) ignores hard queries, (M) includes both easy and hard ones, while (H) considers hard queries only. We report the \gls{map} for the  Medium and Hard splits in all our experiments.

\subsection{Implementation \& Training Details}

\paragraph{Category-level Retrieval.}
The transformer-based models and their pre-trained weights are based on the public implementation\footnote{https://github.com/facebookresearch/deit} of DeiT \cite{touvron2020deit} built upon the Timm library by~\citet{rw2019timm}.  All models are optimized using the AdamW optimizer \cite{loshchilov2017decoupled} with learning rate $3.10^{-5}$, weight decay $5.10^{-4}$ and batch size of 64. 
For all experiments, unless mentioned otherwise, the contrastive loss margin is set to $\beta=0.5$ and the entropy regularization strength is set to $\lambda=0.7$. We show later in ablation that the results are relatively stable (and not overfitted) to this hyper-parameter setting.  
We use standard data augmentation methods of resizing the image to 256$\times$256 and then taking a random crop of size 224$\times$224, combined with random horizontal flipping. Following \citet{wang2020cross}, we use a dynamic offline memory queue of the same size as the dataset (with the exception of In-Shop dataset for which the memory size is 0.2 of the dataset size). Additionally, for the In-Shop dataset we adopt a momentum encoder for the memory entries (similarly to  \citet{he2020momentum}) with momentum value of 0.999. We have found this was not necessary for SOP and CUB-200-2011. Finally, SOP and In-Shop models were trained for 35k iterations and the CUB-200-2011 model was trained for 2000 iterations.
\medskip

\paragraph{Particular Object Retrieval.}
For the particular object retrieval experiments, we build our implementation on top of the public code\footnote{https://github.com/filipradenovic/cnnimageretrieval-pytorch} associated with the work by \citet{radenovic2018fine}. We follow the same optimization and regularization procedure. All models, transformer-based and convolutional, are finetuned using the SFM120k dataset. The input images are resized to have the same fixed width and height. We report results for image sizes of 224$\times$224 and 384$\times$384. For finetuning, each batch consists of 5 tuples of (1 anchor, 1 positive, 5 negatives). For each epoch, we randomly select 2,000 positive pairs and 22,000 negative candidates (using hard-negative mining). We use the default hyper-parameters of \citet{radenovic2018fine}: the models are optimized using Adam \cite{kingma2014adam} with small learning rate of $5.10^{-7}$ and weight decay of $10^{-6}$. The contrastive loss margin  is set to $\beta=0.85$. The models are finetuned for 100 epochs. All models with \gls{gem} pooling use a pooling exponent value of $p=3$. The dimensionality reduction is achieved using a PCA trained on the SFM120k dataset. For the evaluation, all the query and database images are resized into a square image with  the same resolution as used during the finetuning stage.

\begin{table*}[htb]
    \caption{Recall@K performance for the \gls{sop}, CUB-200 and In-Shop category-level datasets compared to the state-of-the-art methods. \\ $\triangleright$128: reduction to 128 components obtained using PCA.}%
    \smallskip
    \smallskip
    \centering
    \scalebox{0.8}{
    \begin{tabular}{lc|c|cccc|cccc|cccc@{\ }}
        \toprule
         \hspace*{\fill} \multirow{2}{*}{Method} \hspace*{\fill} & \multirow{2}{*}{Backbone} & \multirow{2}{*}{\#dims} & \multicolumn{4}{c}{SOP (K)} & \multicolumn{4}{c}{CUB-200 (K)} & \multicolumn{4}{c}{In-Shop (K)}\\
         \cmidrule{4-15}
         & & & 1 & 10 & 100 & 1000  & 1 & 2 & 4 & 8 & 1 & 10 & 20 & 30\\
         \midrule
         A-BIER \cite{opitz2018deep} & \multirow{4}{*}{GoogleNet}
         & \multirow{4}{*}{512} & 74.2 & 86.9 & 94.0 & 97.8 & 57.5 & 68.7 & 78.3 & 86.2 & 83.1 & 95.1 & 96.9 & 97.5 \\
         ABE \cite{kim2018attention} &  &  & 76.3 & 88.4 & 94.8 & 98.2 & 60.6 & 71.5 & 79.8 & 87.4  & 87.3 & 96.7 & 97.9 & 98.2 \\
         SM \cite{suh2019stochastic} &  &  & 75.3 & 87.5 & 93.7 & 97.4 & 56.0 & 68.3 & 78.2 & 86.3  & 90.7 & 97.8 & 98.5 & 98.8 \\
         XBM \cite{wang2020cross}  & &  & 77.4 & 89.6 & 95.4 & 98.4 & 61.9 & 72.9 & 81.2 & 88.6 & 89.4 & 97.5 & 98.3 & 98.6  \\
         \midrule
         HTL \cite{ge2018deep} & 
         \multirow{5}{*}{InceptionBN} &
         \multirow{5}{*}{512} &
         74.8 & 88.3 & 94.8 & 98.4 & 57.1 & 68.8 & 78.7 & 86.5 & 80.9 & 94.3 & 95.8 & 97.2 \\
         MS \cite{wang2019multi} & & & 78.2 & 90.5 & 96.0 & 98.7 & 65.7 & 77.0 & 86.3 & 91.2 & 89.7 & 97.9 & 98.5 & 98.8 \\
         SoftTriple \cite{qian2019softtriple} &  & & 78.6 & 86.6 & 91.8 & 95.4 & 65.4 & 76.4 & 84.5 & 90.4 \\
         XBM \cite{wang2020cross}  & & & 79.5 & 90.8 & 96.1 & 98.7 & 65.8 & 75.9 & 84.0 & 89.9 & 89.9 & 97.6 & 98.4 & 98.6 \\
         HORDE \cite{jacob2019metric} &  &  & 80.1 & 91.3 & 96.2 & 98.7 & 66.8 & 77.4 & 85.1 & 91.0  & 90.4 & 97.8 & 98.4 & 98.7 \\
         \midrule
         Margin \cite{wu2017sampling} & %
         \multirow{8}{*}{ResNet-50}
         & \multirow{4}{*}{128} & 72.7 & 86.2 & 93.8 & 98.0 & 63.9 & 75.3 & 84.4 & 90.6  & - & - & - & - \\
         FastAP \cite{cakir2019deep} &  & & 73.8 & 88.0 & 94.9 & 98.3 & - & - & - & -  & - & - & - & - \\
         MIC \cite{roth2019mic} &  & & 77.2 & 89.4 & 94.6 & - & 66.1 & 76.8 & 85.6 & -  & 88.2 & 97.0 & - & 98.0   \\
         XBM \cite{wang2020cross}  &  & & 80.6 & 91.6 & 96.2 & 98.7 & - & - & - & - & 91.3 & 97.8 & 98.4 & 98.7 \\
         \cmidrule{3-15}
         NSoftmax \cite{zhai2018classification} &  & \multirow{2}{*}{512} & 78.2 & 90.6 & 96.2 & - & 61.3 & 73.9 & 83.5 & 90.0  & 86.6 & 97.5 & 98.4 & 98.8 \\
         ProxyNCA++ \cite{teh2020proxynca} &  &  & 80.7 & 92.0 & 96.7 & 98.9 & 69.0 & 79.8 & 87.3 & 92.7 & 90.4 & 98.1 & 98.8 & 99.0 \\
         \cmidrule{3-15}
         NSoftmax \cite{zhai2018classification} &  & \multirow{2}{*}{2048} & 79.5 & 91.5 & 96.7 & -  & 65.3 & 76.7 & 85.4 & 91.8  & 89.4 & 97.8 & 98.7 & 99.0 \\
         ProxyNCA++ \cite{teh2020proxynca} & & & 81.4 & 92.4 & 96.9 & 99.0 & 72.2 & 82.0 & 89.2 &
         \textbf{93.5}  & 90.9 & \textbf{98.2} &
         \textbf{98.9} & \textbf{99.1} \\
         \midrule
         \multirow{3}{*}{\irtr \textbf{(ours)} } &
         \multirow{2}{*}{DeiT-S}
         & $\triangleright$128 & 83.4 & 93.0 &
         97.0 & 99.0 &  72.6 & 81.9 &
         88.7 & 92.8 & 91.1 & 98.1 &
         98.6 & 99.0 \\
         &  & \pzo384 & \textbf{84.0}
         & \textbf{93.6} & \textbf{97.2} & \textbf{99.1}  & \textbf{74.7}
         & \textbf{82.9} & \textbf{89.3} & 93.3 & \textbf{91.5}
         & 98.1 & 98.7 & 99.0 \\
         & DeiT-S$\dagger$ & \pzo384 &
         \textbf{84.2} & \textbf{93.7} & \textbf{97.3} & \textbf{99.1} & \textbf{76.6} & \textbf{85.0} & \textbf{91.1} & \textbf{94.3}  &\textbf{91.9} & 98.1 & 98.7 & 98.9  \\
         \bottomrule
    \end{tabular}}
    \label{tab:main_results_category}
    \vspace*{-0.5\baselineskip}
\end{table*}

\subsection{Results}

\paragraph{Category-level Retrieval.} We present the Recall@K performance for three public benchmarks for category-level retrieval. For the SOP dataset, we can see in Table~\ref{tab:main_results_category} that our \irtr model with DeiT-S$^{384}$ backbone achieves state-of-the-art performance for all values of K, outperforming previous methods by a margin of 2.6\% absolute points for Recall@1. The DeiT-S$\dagger$ variant with distillation pre-training achieves the best results on this benchmark. Even when reducing the dimensionality to 128-D, our method outperforms all the convnets except at  Recall@1000. 
On the CUB-200-2011 dataset, %
the DeiT-S$^{384}$ model outperforms the current state of the art by 2.5\% points at Recall@1.  The distilled DeiT-S model provides an additional 1.9\% improvement, achieving the best results for all values of K. The {DeiT-S$^{128}$} variant with compressed representation outperforms all previous methods except for the ProxyNCA++ model that uses 2048-D descriptors. Similarly, for In-Shop, the DeiT-S$^{384}$ model and its distilled variant outperform all previous models at Recall@1 with a margin of 0.2\% and 0.6\% respectively. 

\paragraph{Particular Object Retrieval.} We present the \gls{map} performance for the Medium and Hard splits of the revisited Oxford and Paris benchmarks in Table~\ref{tab:trad_ret}. First observe that for input images with size 224$\times$224, the DeiT-S$\dagger$ backbone outperforms its ResNet-50 counterpart with the same capacity, as well as the higher capacity ResNet-101 across all benchmarks and descriptor sizes. The larger DeiT-B$\dagger$ provides a significant gain in performance and achieves the best result among the reported models. Scaling up the image size to 384$\times$384 considerably improves the performance for all models with the DeiT-B$\dagger$ model retaining its position as the strongest model.  %
In Table~\ref{tab:trad_ret_sota} we compare our model to strong state-of-the-art methods in particular object retrieval, following the standard extensive evaluation procedure. \citet{revaud2019learning} use the original resolution of the dataset (i.e. 1024$\times$768), while \citet{radenovic2018revisiting} utilizes multi-scale evaluation. Although these methods outperform our DeiT-B$\dagger$ model at resolution $384\times384$ in \gls{map}, they are approximately 248\% and 437\% more expensive w.r.t. FLOPS. Furthermore, we observe that for compressed representations of 128-D, our model closes the gap with \citet{radenovic2018revisiting}, achieving a higher \gls{map} for $\mathcal{R}$Paris.

\begin{table}[t]\phantom{o}\vspace*{-28pt}\\
    \caption{Ablation of model components: off-the-shelf performance  (\irto), with contrastive learning (\irtf) and finally regularized (\irtr). All methods use a DeiT-S backbone with \#dims=384.
    \label{tab:irt_compare}}
    \smallskip 
    \centering 
    \scalebox{0.8}{
        \begin{tabular}{c|ccc|cc|cc}
        \toprule
         \hspace*{\fill}\multirow{2}{*}{Supervision $\downarrow$\!}\hspace*{\fill}& \multirow{2}{*}{SOP} & \multirow{2}{*}{CUB} & \!\multirow{2}{*}{In-Shop} & \multicolumn{2}{c}{$\mathcal{R}$Ox} & \multicolumn{2}{c}{$\mathcal{R}$Par} \\
         \cmidrule{5-6}
         \cmidrule{7-8}
          & & & & M & H & M & H \\
         \midrule
          \irto &  52.8 & 58.5 & 31.3 & 20.3 & 5.6 & 50.2 & 26.3 \\
           \irtf  & 83.0 & 74.2 & 90.3 &  32.7 & 11.4 &  63.6 & 37.8   \\
           \irtr & \textbf{84.0} & \textbf{74.7} & \textbf{91.5} & \textbf{34.0} & \textbf{11.5} & \textbf{66.1} & \textbf{40.2} \\
         \bottomrule
    \end{tabular}}
 \vspace*{-10pt}
\end{table}

\begin{table*}[t!]
    \begin{minipage}[t]{0.28\textwidth}
    \caption{Comparison between convolutional ResNet-50 (R50) and \irtf DeiT-Small (DeiT-S) architectures across multiple metric learning objective functions, as tested using the SOP dataset, $\beta=0.5$ ("Contr." refers to the contrastive loss we use).}\vspace*{6pt}%
    \centering
    \scalebox{0.8}{
        \begin{tabular}{lcc|c}
        \toprule
         \hspace*{\fill}Model\hspace*{\fill} & \hspace*{\fill}Loss\hspace*{\fill} & \#dims & R@1 \\
         \midrule
         R50 %
         & \!\!\multirow{2}{*}{NSoftmax}\!\! & 2048 & 79.5 \\
         DeiT-S &  & \pzo384 &  \textbf{80.8} \\
         \midrule
         R50 & \multirow{2}{*}{SNCA} & 2048 & 78.0 \\
         DeiT-S & & \pzo384 & \textbf{81.1}  \\
         \midrule
         R50 & \multirow{2}{*}{Contr.} & 2048 & 79.8 \\
         DeiT-S\;\;\; & & \pzo384 & \textbf{83.0} \\
         \bottomrule
    \end{tabular}}
    \label{tab:loss_fns}
    \end{minipage}\hspace*{\fill}
    \begin{minipage}[t]{0.35\textwidth}
    \caption{Recall@1 results for different values of margins $\beta$ and entropy regularization strengths $\lambda$ on SOP.
    Entropy regularization consistently boosts
    performance (drops again for $\lambda>1.0$).}\vspace*{6.5pt}%
    \begin{small}
    \footnotesize
    \centering
    \scalebox{0.8}{
    \centering
        \begin{tabular}{l|c|ccccc}
        \toprule
         \hspace*{\fill}\multirow{2}{*}{Model}\hspace*{\fill} & \hspace*{\fill}\multirow{2}{*}{\;\;$\beta$\;\;}\hspace*{\fill} & \multicolumn{5}{c}{ $\lambda$} \\
   		\cmidrule(r){3-7} & & 0.0 & 0.3 & 0.5 & 0.7 & 1.0  \vspace*{1pt}\\
         \midrule
          & 0.1  & 78.9 & \textbf{79.2} & 79.1 & 78.8 & 78.8  \\
          & 0.3  & 79.3 & 81.3 & \textbf{81.7} & \textbf{81.7} & \textbf{81.7}  \\
         R50 & 0.5  & 79.8 & 80.7 & 81.0 & 81.2 & \textbf{81.3}  \\
          & 0.7  & 79.3 & 80.5 & \textbf{81.2} & 81.1 & 78.5  \\
          & 0.9  & 79.1 &\textbf{ 80.0} & 31.4 & 13.4 & 9.1  \\[1.31pt]
          \midrule
          & 0.1  & 70.3  & 70.7 &\textbf{ 71.4} & 70.1 & 71.1   \\
          & 0.3  &  82.5 & 83.0 & 83.0 & \textbf{83.1} & \textbf{83.1}  \\
          DeiT-S\! & 0.5  & 83.0 & 83.5 & 83.6 & \textbf{84.0} & \textbf{84.0}  \\
          & 0.7  & 82.9 & 83.8 & 84.1 & \textbf{84.2} &  83.8 \\
          & 0.9  & 82.6 & \textbf{84.4} & 80.6 & 71.2 &  41.5 \\[2pt]
          \bottomrule
    \end{tabular}
    }
    \end{small}
    \label{tab:la_beta_grid}
    \end{minipage}\hspace*{\fill}%
    \begin{minipage}[t]{0.31\textwidth}
    \caption{Particular object retrieval \gls{map} performance for different entropy regularization strengths $\lambda$. The results are obtained by using DeiT-Small model with CLS token as the feature descriptor (\#dims=384, trained with contrastive loss).}\vspace*{7.5pt}%
    \centering
    \scalebox{0.8}{
        \begin{tabular}{l|c|cc|cc}
        \toprule
        \multirow{2}{*}{Model} &
        \multirow{2}{*}{$\lambda$}
        & \multicolumn{2}{c}{$\mathcal{R}$O} & \multicolumn{2}{c}{$\mathcal{R}$Par} \\ %
        \cmidrule{3-4}
        \cmidrule{5-6}
          &  & M & H & M & H \\ %
         \midrule
         \multirow{6}{*}{DeiT-S} & \multirow{1}{*}{0.0} &   32.7 & 11.4 &  63.6 & 37.8 \\ %
         \cmidrule{2-6}
            & \multirow{1}{*}{1.0} &  31.5 & 9.3 & 64.7 & 38.6  \\ %
            & \multirow{1}{*}{2.0}   &34.5 & 11.1 & 65.7 & 39.8  \\ %
            & \multirow{1}{*}{3.0} &  \textbf{34.6} &\textbf{11.5} & \textbf{66.1} & 40.1  \\ %
            & \multirow{1}{*}{4.0} &  34.0 & \textbf{11.5} & \textbf{66.1} & \textbf{40.2}  \\ %
            & \multirow{1}{*}{5.0} &  32.3 & 10.4 & 65.6 & 39.8 \\[0.2pt] %
         \bottomrule
    \end{tabular}}
    \label{tab:trad_ret_entropy}
    \end{minipage}
    \vspace*{-16pt}
\end{table*}

\begin{table}[h!]
    \caption{Particular object retrieval \gls{map} performance comparison between different convolutional and
    \irtf models using different descriptor dimensions. All models are finetuned the same way. \\ $\triangleright$128: reduction to 128 components obtained using PCA.%
    \label{tab:trad_ret}}
    \smallskip
    \centering 
    \scalebox{0.8}{%
        \begin{tabular}{llcc |cc|cc}
        \toprule
        \multirow{2}{*}{Input size} & \multirow{2}{*}{Model} & \multirow{2}{*}{Pooler}& \multirow{2}{*}{\#dims} & \multicolumn{2}{c}{$\mathcal{R}$Ox} & \multicolumn{2}{c}{$\mathcal{R}$Par} \\
        \cmidrule{5-6}
        \cmidrule{7-8}
          & & & & M & H & M & H \\
         \midrule
         \multirow{6}{*}{224$\times$224} & R50 & GeM & \multirow{6}{*}{$\triangleright$128} & 25.8 & 8.6 & 56.7 & 31.2 \\
         &R50 & R-MAC &  &   23.6 & 5.5 &  56.0 & 30.8 \\
         &R101 & GeM &  & 27.8 & 8.0 & 59.0 & 32.2 \\
         &R101 & R-MAC &  & 27.3 & 7.4 & 57.9 & 31.3  \\
         &DeiT-S$\dagger$\!\! & CLS &  & 32.1 & 13.3 & 63.8  & \textbf{39.3} \\
         &DeiT-B$\dagger$\!\! & CLS &  & \textbf{36.6} & \textbf{14.8} & \textbf{64.4} & 39.1  \\
         \midrule
         \multirow{6}{*}{224$\times$224} &R50 & GeM & \multirow{4}{*}{2048} &  28.7 & 10.9 & 61.2 & 35.9  \\
         &R50 & R-MAC &  &  25.6 & 7.3 &  60.6 & 35.4 \\
         &R101 & GeM &  &  31.7 & 11.1 & 63.4 & 37.3 \\
         &R101 & R-MAC & & 31.0 & 9.3 & 62.6 & 36.5  \\
         &DeiT-S$\dagger$\!\! & CLS & \pzo384 &  34.5 & 15.8 &  65.8 & 42.0 \\
         &DeiT-B$\dagger$\!\! & CLS & \pzo768 & \textbf{39.5} & \textbf{17.4} & \textbf{67.5}  & \textbf{43.6}  \\
        \toprule
         \multirow{3}{*}{384$\times$384} & R101 & GeM & \multirow{3}{*}{$\triangleright$128}  & 34.1 & 9.5  & 62.6 & 36.3 \\
         &R101 & R-MAC &   & 31.4 & 7.4  & 61.6 & 35.3  \\
         &DeiT-B$\dagger$\!\! & CLS &   & \textbf{49.0} & \textbf{21.5}  & \textbf{68.5} & \textbf{43.8} \\
         \midrule
         \multirow{3}{*}{384$\times$384} &R101 & GeM & \multirow{2}{*}{2048}  & 38.1 & 12.5  & 69.4 & 45.8 \\
         &R101 & R-MAC &   & 37.1 & 10.6  & 66.0 & 41.4  \\
         &DeiT-B$\dagger$\!\! & CLS & \pzo768  & \textbf{50.5} & \textbf{22.7}  & \textbf{70.6} & \textbf{47.4} \\[2.3pt]
         \bottomrule
     \end{tabular}
    }
\end{table}

\begin{table}[htb]
    \begin{minipage}[t]{\linewidth}
    \caption{Comparison with SoA methods for particular object retrieval: [1] \cite{radenovic2018revisiting}, [2]  \cite{revaud2019learning}. 
    \\ $\triangleright$128: reduction to 128 components obtained using PCA. \\
    $\star$: FLOPS (G) are computed for input images of size 1024$\times$768.  \\
    $\S$: our evaluation 
    using pre-trained models from the authors.
    }\smallskip
    \centering
    \scalebox{0.75}{
        \begin{tabular}{lc|c@{\ }|@{\ }c@{\ }|cc|cc}
        \toprule
        \hspace*{\fill}\multirow{2}{*}{Method}\hspace*{\fill}& \!\!\!\!\!\!\!\multirow{2}{*}{
        \makebox{\begin{minipage}{1.0cm}\ \hspace*{6pt}Model \{maxres\}\end{minipage}}
            } & \multirow{2}{*}{\#dims\ } & \multirow{2}{*}{\makebox{\begin{minipage}{1cm}\ \hspace*{0pt}FLOPS \hspace*{7pt}(G)\end{minipage}}}& \multicolumn{2}{c}{$\mathcal{R}$Ox} & \multicolumn{2}{c}{$\mathcal{R}$Par} \\
        \cmidrule{5-6}
        \cmidrule{7-8}
         &&& & M & H  & M & H \\
         \midrule
         \irtf & \!\!\!\!\multirow{4}{*}{DeiT-B$\dagger$\{384\}}\!  & \multirow{2}{*}{$\triangleright$128} &  \multirow{4}{*}{98.8} & 49.0 & 21.5  & 68.5 & 43.8 \\
         \irtr &   &  &  & 49.1 & 21.1 & 68.3 & 44.1 \\
         \irtf &    & \multirow{2}{*}{\pzo768} &   & 50.5 & 22.7  & 70.6 & 47.4 \\
         \irtr &   &  &  & 55.1 & 28.3 & 72.7 & 49.6 \\
         \midrule
         $[1]$-GeM$\S$ & \!\!\!\multirow{2}{*}{R101\{1024\}} & $\triangleright$128 & \multirow{2}{*}{432.2$\star$}  & 53.2 & 28.9  & 65.4 & 36.9  \\
         $[1]$-GeM & & 2048 &  & 64.7 & 38.5  & 77.2 & 56.3  \\
         \midrule
         $[2]$-GeM & \!\!\!R101\{1024\} & 2048 & 246.0$\star$ & 67.2 & 42.8 & 80.1 & 60.5 \\ 
         \bottomrule
    \end{tabular}}
    \label{tab:trad_ret_sota}
    \end{minipage}\\
    \begin{minipage}[t]{\linewidth}
    \caption{Performance of different pooling methods on both retrieval tasks %
    (\irtf model, DeiT-Small backbone, \#dims=384).}%
    \vspace*{5pt}
    \scalebox{0.8}{
        \begin{tabular}{l|ccc|cc|cc}
        \toprule
         \hspace*{\fill}\multirow{2}{*}{Pooler}\hspace*{\fill}& \multirow{2}{*}{SOP} & \multirow{2}{*}{CUB} & \!\!\multirow{2}{*}{In-Shop} & \multicolumn{2}{c}{$\mathcal{R}$Ox} & \multicolumn{2}{c}{$\mathcal{R}$Par} \\
         \cmidrule{5-6}
         \cmidrule{7-8}
          & & & & M & H & M & H \\
         \midrule
         Average Pool  & \textbf{83.0} & 72.8 &  90.2 & 28.3 & \pzo8.5 &  61.9 & 36.0 \\
         Max Pool  & 82.2 & 69.2 & 90.3 & 25.2 & \pzo6.8 & 60.4 & 34.1 \\
         GeM  & 82.6 & 69.1 & 89.8& 26.5 & \pzo8.5 & 60.2 & 33.7   \\
         CLS  & \textbf{83.0} & \textbf{74.4} & \textbf{90.4} & \textbf{32.7} & \textbf{11.4} & \textbf{63.6}& \textbf{37.8} \\

         \bottomrule
    \end{tabular}}
    \label{tab:pooling}
    \end{minipage}
\end{table}

\subsection{Ablations}

\paragraph{Different Methods of Supervision.}
We provide a comparison between different degrees of supervision corresponding to \irto, \irtf and \irtr in Table~\ref{tab:irt_compare}. We observe that finetuning substantially improves performance over off-the-shelf features, especially for category-level retrieval. Augmenting the contrastive loss with differential entropy regularization further improves the performance across all benchmarks. Figure~\ref{fig:hist_deit} demonstrates how the distribution of the cosine similarities between positive and negative pairs is impacted by the different variants we study. We notice that finetuning strongly helps to make the positive and negative distributions more separable. The entropy regularization term spreads the similarity values across a wider range. %

\paragraph{Choice of Feature Extractor: Pooling Methods.} 

In Table~\ref{tab:pooling}, we study different feature aggregation methods, as described in Section~\ref{sec:off-shelf}. Both for category-level and particular object retrieval, we observe that utilizing the CLS token as the image-level descriptor provides the strongest performance (or at least on par) compared to other popular pooling methods such as average pooling, max pooling and \gls{gem}. This suggests that the transformer operates as a learned aggregation operator, thereby %
reducing the need for careful design of feature aggregation methods. %

\paragraph{Performance across Objective Functions.}  The choice of the objective function used to train image descriptors is crucially important
and is the focus of the majority of the metric learning research. While we adopt the contrastive loss as our primary objective function, we additionally investigate two objective functions
with different properties: (1) Normalized Softmax \cite{zhai2018classification} as a classification-based objective, and (2) Scalable \gls{nca} \cite{wu2018improving}, a pairwise objective with implicit
weighting of hard negatives through temperature.
Table~\ref{tab:loss_fns} shows that DeiT-S outperforms its convolutional
counterpart across all different choices of objective functions. This suggests that 
transformer-based models are strong metric learners and hence an attractive alternative to
convolutional models for image retrieval.

\paragraph{Regularizing Hyper-parameter $\lambda$.} We explore the differential entropy regularization strength and its impact on the improvement of retrieval performance. First, we use the
\gls{sop} dataset for our analysis and show how the Recall@1 performance changes with different
margin values $\beta$ and entropy regularization strengths $\lambda$ in Table~\ref{tab:la_beta_grid}. 

All models, either transformer-based or convolutional, trained
with different margins are improved by the $\lentropy$ regularizer.
The margins with the best results are those with the lowest $\gamma$ values in
Figure~\ref{fig:nuclear_cov}.  Moreover, we observe a similar boost in performance for particular object retrieval in Table~\ref{tab:trad_ret_entropy}, confirming that 
the differential entropy regularization provides a clear and consistent improvement across different tasks and architectures. %

\ignore{
\begin{table}[htb]
    \caption{[OBSOLETE] Comparison between ResNet-50 (R50) and transformer-based DeiT-Small (DeiT-S) architecture across multiple metric learning objective functions, as tested using the SOP dataset.}
    \begin{center}
    \scalebox{0.8}{
        \begin{tabular}{lcc|c}
        \toprule
         \hspace*{\fill}Loss\hspace*{\fill}\hspace*{\fill} & Backbone & \# dims & R@1 \\
         \midrule
         \multirow{2}{*}{NSoftmax} %
         & R50 & 2048 & 79.5 \\
        & DeiT-S & \pzo384 &  \textbf{80.8} \\
         \midrule
         \multirow{2}{*}{SNCA} & R50 & 2048 & 78.0 \\
         & DeiT-S & \pzo384 & \textbf{81.1}  \\
         \midrule
         \multirow{2}{*}{Contrastive} & R50 & 2048 & 79.8 \\
         & DeiT-S & \pzo384 & \textbf{83.0} \\
         \bottomrule
    \end{tabular}}
    \end{center}
    \label{tab:loss_fns}
    \vspace*{-\baselineskip}
\end{table}

\begin{table}[htb]
    \caption{[OBSOLETE] Recall@1 results for different values of margins $\beta$ and entropy regularization strengths $\lambda$
    reported using the \gls{sop} dataset. We show results for both ResNet-50 and 
    transformer-based DeiT-Small and demonstrate that entropy regularization consistently boosts
    the performance across all margins and architectures. Performance with larger $\lambda$ values is illustrated Figure~\ref{fig:r@1_la}.}
    \begin{center}
    \begin{small}
    \footnotesize
    \scalebox{0.85}{
        \begin{tabular}{l|c|ccccc}
        \toprule
         \hspace*{\fill}\multirow{2}{*}{Backbone}\hspace*{\fill} & \hspace*{\fill}\multirow{2}{*}{\;\;$\beta$\;\;}\hspace*{\fill} & \multicolumn{5}{c}{ $\lambda$} \\
   		\cmidrule(r){3-7} & & 0.0 & 0.3 & 0.5 & 0.7 & 1.0  \\
         \midrule
          & 0.1  & 78.9 & \textbf{79.2} & 79.1 & 78.8 & 78.8  \\
          & 0.3  & 79.3 & 81.3 & \textbf{81.7} & \textbf{81.7} & \textbf{81.7}  \\
         R50 & 0.5  & 79.8 & 80.7 & 81.0 & 81.2 & \textbf{81.3}  \\
          & 0.7  & 79.3 & 80.5 & \textbf{81.2} & 81.1 & 78.5  \\
          & 0.9  & 79.1 &\textbf{ 80.0} & 31.4 & 13.4 & 9.1  \\
          \midrule
          & 0.1  & 70.3  & 70.7 &\textbf{ 71.4} & 70.1 & 71.1   \\
          & 0.3  &  82.5 & 83.0 & 83.0 & \textbf{83.1} & \textbf{83.1}  \\
          DeiT-S\! & 0.5  & 83.0 & 83.5 & 83.6 & \textbf{84.0} & \textbf{84.0}  \\
          & 0.7  & 82.9 & 83.8 & 84.1 & \textbf{84.2} &  83.8 \\
          & 0.9  & 82.6 & \textbf{84.4} & 80.6 & 71.2 &  41.5 \\
          \bottomrule
    \end{tabular}
    }
    \end{small}
    \end{center}
    \label{tab:la_beta_grid}
    \vspace*{-\baselineskip}
\end{table}

\begin{table}[htb]
    \caption{[OBSOLETE] Particular object retrieval \gls{map} and globalAP performance for different entropy regularization strengths $\lambda$. The results are using DeiT-Small model with CLS token as the feature descriptor.}
    \begin{center}
    \scalebox{0.8}{
        \begin{tabular}{l|cc|cc|cc|cc}
        \toprule
        & \multicolumn{4}{c}{\gls{map}} & \multicolumn{4}{|c}{globalAP} \\
        \cmidrule{2-5}
        \cmidrule{6-9}
        & \multicolumn{2}{c}{$\mathcal{R}$O} & \multicolumn{2}{c}{$\mathcal{R}$Par} & \multicolumn{2}{|c}{$\mathcal{R}$O} & \multicolumn{2}{c}{$\mathcal{R}$Par} \\
        \cmidrule{2-3}
        \cmidrule{4-5}
        \cmidrule{6-7}
        \cmidrule{8-9}
           $\lambda$ & M & H & M & H &  M & H & M & H \\
         \midrule
         \multirow{1}{*}{0.0} &   32.7 & 11.4 &  63.6 & 37.8 & 31.9 & 3.1 & 61.0 & 33.8 \\   
         \midrule
             \multirow{1}{*}{1.0} &  31.5 & 9.3 & 64.7 & 38.6 & 32.5 & 3.0 & 61.5 & 34.4 \\
             \multirow{1}{*}{2.0}   &34.5 & 11.1 & 65.7 & 39.8 & 36.4 & 3.7 & 62.4 & 35.1 \\
             \multirow{1}{*}{3.0} &  \textbf{34.6} &\textbf{11.5} & \textbf{66.1} & 40.1 & 39.7 & 4.5 & 63.4 & 35.8 \\
             \multirow{1}{*}{4.0} &  34.0 & \textbf{11.5} & \textbf{66.1} & \textbf{40.2} & \textbf{41.3} & \textbf{4.9} & \textbf{63.5} & \textbf{35.9} \\
             \multirow{1}{*}{5.0} &  32.3 & 10.4 & 65.6 & 39.8 & 40.7 & 4.8 & 63.1 & 35.5\\
         \bottomrule
    \end{tabular}}
    \end{center}
    \label{tab:trad_ret_entropy}
    \vspace*{-\baselineskip}
\end{table}
} 

\section{Conclusion}

In this paper,  we have explored how to adapt the transformer architecture to metric learning and image retrieval. 
In this context, we have revisited the contrastive loss formulation and showed that a regularizer based on a differential entropy loss spreading vectors over the unit hyper-sphere improves the performance for transformer-based models, as well as for convolutional  models.
As a result, we establish the new state of the art for category-level image retrieval. Finally, we demonstrated  that, for comparable settings, transformer-based models are an attractive alternative to convolutional backbones for particular object retrieval, especially with short vector representations. Their performance is competitive against convnets having a much higher complexity.

\begingroup
    \setlength{\bibsep}{4.2pt}
    \bibliography{egbib}
    \bibliographystyle{icml2021}
\endgroup

\end{document}